\documentclass[runningheads]{llncs}
\usepackage{graphicx}
\usepackage{xcolor}
\usepackage{multirow}
\usepackage{float}
\usepackage{booktabs}
\usepackage{color}
\usepackage{colortbl}
\usepackage{tabularx} 
\usepackage{tabulary} 
\begin{document}
\title{Denoising Opponents Position in Partial Observation Environment}
\titlerunning{Denoising in Partial Observation Environment}
%
\author{
Aref Sayareh\inst{1}\and
Aria Sardari\inst{1}\and
Vahid Khoddami\inst{1}\and
Nader Zare\inst{2}\and 
Vinicius Prado da Fonseca\inst{1}\and 
Amilcar Soares\inst{1}}
\authorrunning{A. Sayareh et al.}
%
\institute{
Memorial University of Newfoundland, St. John's, Canada\\
\and
Institute for Big Data Analytics, Dalhousie University, Halifax, Canada\\
\email{\{asayareh, asardariloud, vkhoddami, vpradodafons, asoaresjunio\}@mun.ca},\\
\email{nader.zare@dal.ca},\\ 
}
\maketitle              
\begin{abstract}
The RoboCup competitions hold various leagues, and the Soccer Simulation 2D League is a major among them. Soccer Simulation 2D (SS2D) match involves two teams, including 11 players and a coach for each team, competing against each other. The players can only communicate with the Soccer Simulation Server during the game. 
Several code bases are released publicly to simplify team development. 
So researchers can easily focus on decision-making and implementing machine learning methods. 
SS2D actions and behaviors are only partially accurate due to different challenges, such as noise and partial observation.
Therefore, one strategy is to implement alternative denoising methods to tackle observation inaccuracy.
Our idea is to predict opponent positions while they have yet to be seen in a finite number of cycles using machine learning methods to make more accurate actions such as pass.
We will explain our position prediction idea powered by Long Short-Term Memory models (LSTM) and Deep Neural Networks (DNN).
The results show that the LSTM and DNN predict the opponents' position more accurately than the standard algorithm, such as the last-seen method.


\end{abstract}
\section{Introduction}
RoboCup has hosted annual robotic soccer competitions since 1997, following the proposal of robotic soccer games as a new research topic in 1992 \cite{robo1997,noda1996soccer,kitano1997robocup}. 
These competitions aim to advance the robotics and artificial intelligence field by implementing robotic soccer games.
Several leagues have been created to achieve this, including rescue, soccer simulation, and standard platform soccer.
  
The Soccer Simulation 2D (SS2D) league is one of the oldest RoboCup leagues.
In a soccer simulation 2D game, two teams containing 11 players and one coach compete against each other. 
Each game takes 6000 cycles (10 minutes). 
In each cycle, the players and coach receive observations from the RoboCup Soccer Simulation Server and send actions based on those observations back to the server.
The server is responsible for controlling the game.

\subsection{Sensors}
Each player can update their observation of the field using three sensors.
The sensors are described shortly below:

\begin{itemize}
  \item \textbf{Visual Sensor}: This sensor lets agents see the fields in three view area modes. However, as the view area increases, this sensor receives more time-out.
  \item \textbf{Body Sensor}: Each player will receive information about their physical body in the field using Body Sensor every cycle. This information contains the player's speed, stamina, view mode, and neck angle.
  \item \textbf{Hear Sensor}: Players can communicate with each other by saying and hearing 10 characters and only listening to only one player per cycle. 
\end{itemize}

\subsection{Challenges in Visual Sensor}

\begin{figure}[!ht]
\centering
\includegraphics[width=0.9\textwidth]{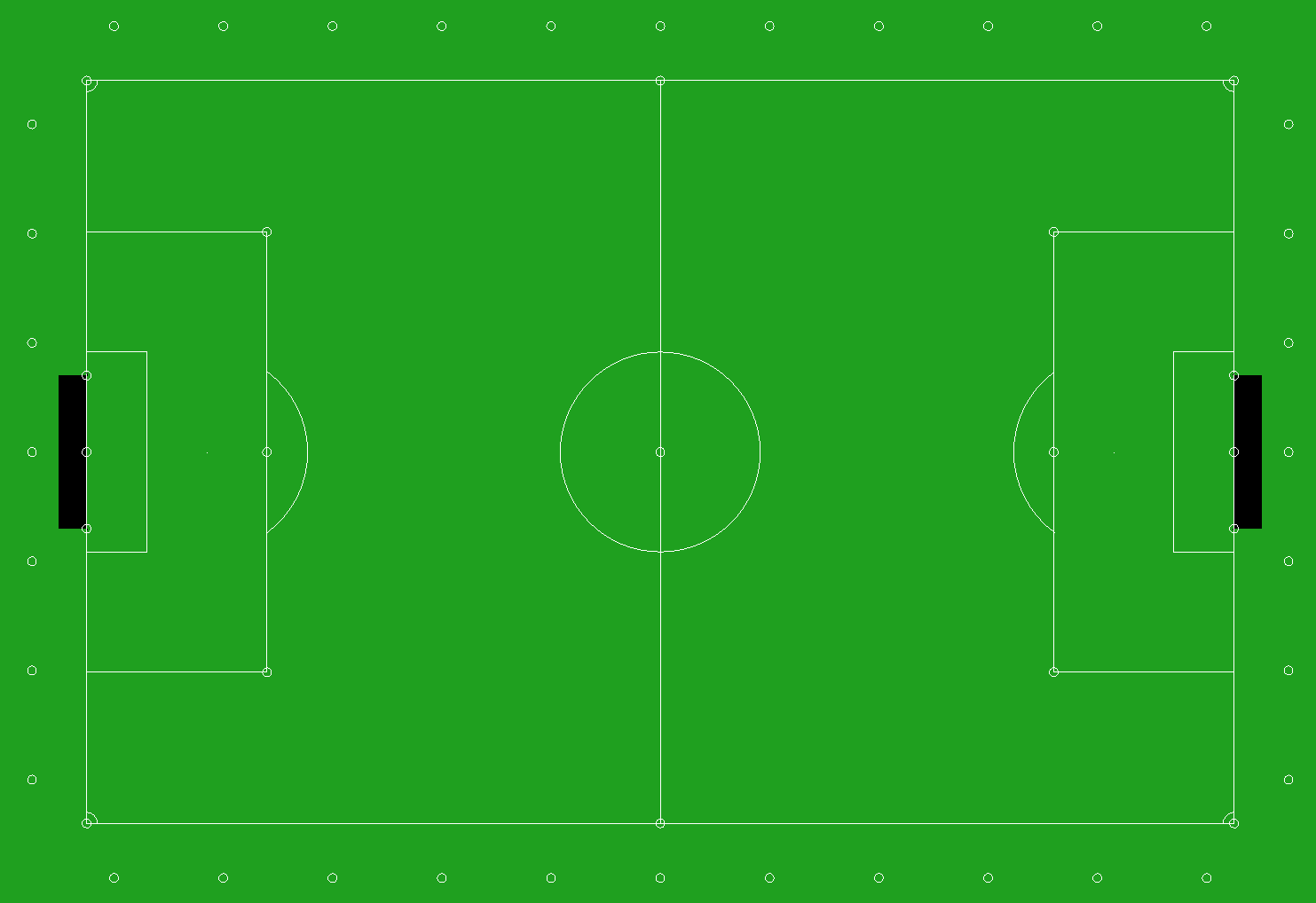}
\caption{White circles are the flags in the field.}
\label{flgs}
\end{figure}

An agent must first find its location in the field, then find other players and the ball's location relative to itself.
Fig.~\ref{flgs} showed several flags considered objects, such as the posts of goals, the corner of the fields, and some points on the lines of the soccer field.
As mentioned, the visual sensor receives relative coordination of all objects in the view area, which contains the object's distance to the agent and the direction from its neck. 

The game server adds some noise to the distance and the direction of each coordination To make the environment more challenging and closer to reality.
The directions range from -180 to +180 degrees, and their values are rounded with no floating points. 
The distance noise has a quantized formula that will increase based on the distance.
So, as an object's distance grows, the noise of the distance increases.
Therefore, calculating the agent's position (Localization Algorithm) considers the three closest flags in the view area.
First, the agent finds its global neck angle and then calculates its position by averaging the estimated positions with those flags. 

After self-position, the agent estimates other players and the ball's positions if they are in the view area.
The positions of the players and the balls are less accurate than the self-position because there is only one estimation for each object. At the same time, there are at least three flags for self-position estimation. 

Another challenge in the visual sensor is partial observation, in which an agent can not see the whole field in only one cycle.
It takes at least six cycles for an agent to check the entire environment with any View Width.
Thus, the agent must choose a good angle for neck direction to see important objects. 
For example, in offensive situations, the players need to keep track of the ball and the opponent's defenders not to lose the ball in pass or dribble action.
  
\begin{figure}[!ht]
    \centering
    \includegraphics[width=0.48\textwidth]{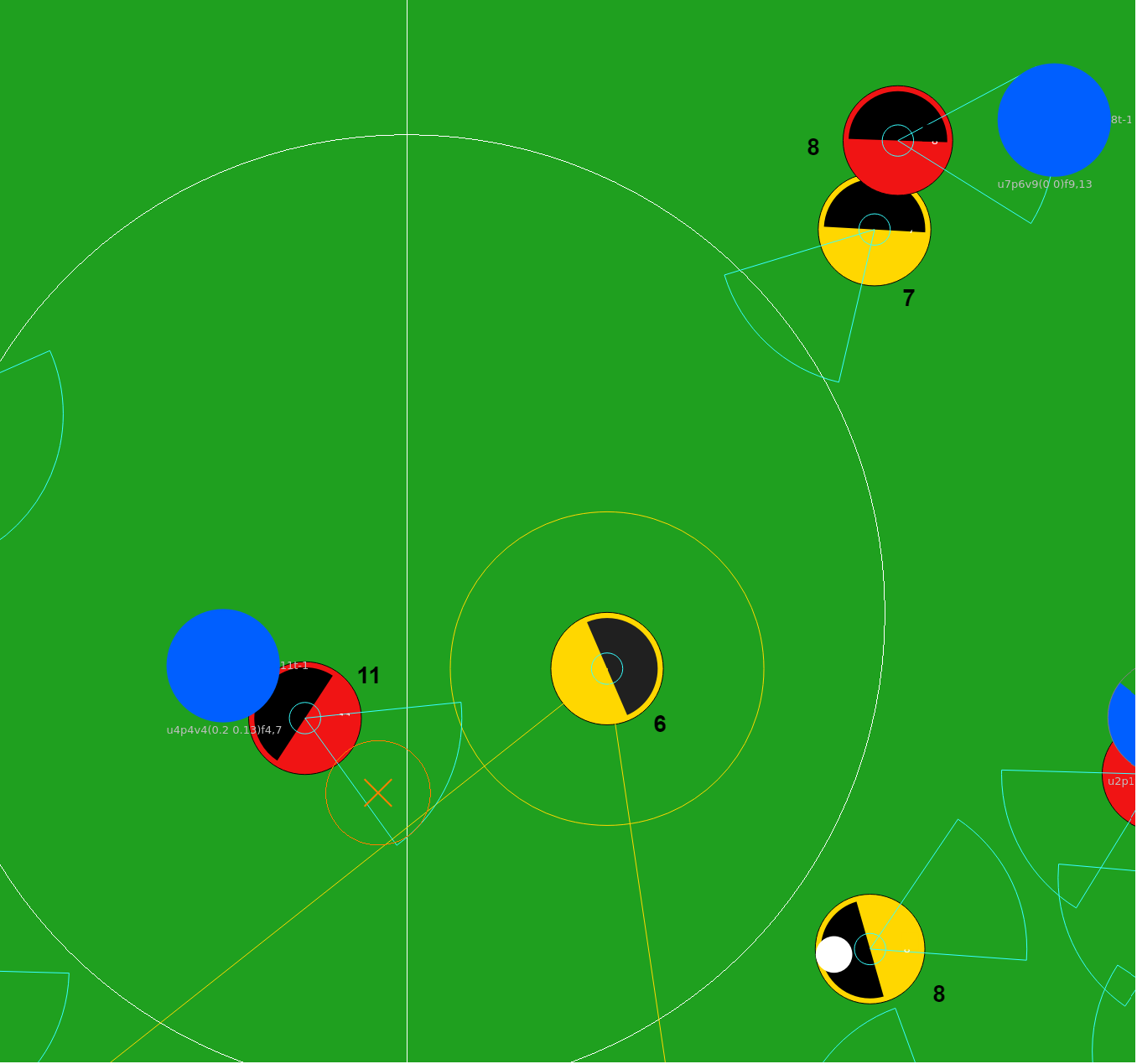}
    \caption{This Figure shows the state that player six of the left team observed. Yellow circles are left team players, red ones are right team players, and the ball is a small white circle. Also, the blue circles are the last position of the opponent players.}
    \label{poscount}
  \end{figure}

In the HeliosBase \cite{heliosbase}, a popular team base among researchers in this field, when the agent receives data about an object using a hearing or visual sensor, a counter will reset to zero.
This counter increases each cycle until the agent receives new data about that object. 
This counter is called Pos Count, which shows how many cycles have passed since the last position update for that player.
Fig.~ \ref{poscount} shows objects considered in their last updated position. 
  
In this paper, we propose a machine-learning approach to predict the players' positions of the opponent team that are not seen in previous cycles.
The prediction method can use only the current state of the field or several previous states of the field to predict more accurately.

In the present paper, Section 2, we discuss some related papers that led us to implement the denoising system. After that, we explain the methods that are used in this project, such as data generation and model selection. Finally, we explained how to evaluate the presented method to reduce the noise in this environment.

\section{Related Works}

Cyrus' team has done feature importance analysis on data for pass behavior prediction \cite{cyruspaper}.
They developed a model to predict the opponent's pass behavior, so they improved their defense behavior in this environment. 
Using their model, they could accurately predict the two players that might receive the pass.
Therefore, their defense players mark those players to block any pass in the future. 
In their paper, the most important features are the position, velocity, and body angle of the players and the position and velocity of the ball.
Therefore, we used those features in our method as the input of our models.

The FraUnited team developed a model to predict the offensive behavior of an opponent team\cite{frapaper}.
The authors analyze their opponent's behavior through the beginning of the game (around 1000 cycles). 
Then, they update their pre-trained model to predict exactly the position of each opponent player.

Nakade et al. employed SIRMs fuzzy models to predict the position of the players when the opponent team is attacking \cite{heliospaper}. 
They used this model in defensive situations to intercept the ball effectively.



\section{Methods}

To create a model to predict the players' position, considerable data is required. Therefore, we generate the data and then modify the unknown objects' data before normalizing it. 
After that, different models are fitted by the data to find the best one to predict the players' positions.

\subsection{Data Generation}
We need a huge amount of data to feed the models to reach a high prediction accuracy. Therefore, we conduct 1000 games between CyrusBase \cite{cyrusbase} and HeliosBase \cite{heliosbase} using AutoTest tool\footnote{AutoTest tool link: https://github.com/Cyrus2D/AutoTest2D.}.
In the decision progress of player 9 of the CyrusBase, a DataExtractor object was added, which extracts all of the information of the objects in the field in each cycle and saves them in CSV format in a file. 

Different files were created due to games running in parallel.
Therefore, in each game, player 9 created a file based on the time the game started and the port number by which the player connected to the server.

\subsection{Data explanation}

After data generation, there were 1000 files for each game that player number 9 on the right side created to save the information of the field.
Each file contains about 4,000 rows, including the cycle game number and noisy data of the ball and 22 players. 
In addition, each sample also has accurate field data, meaning that the exact position of each object in the field is saved in the sample (the pos count of the objects is 0).

Although each game is 6000 cycles, each file has around 4000 cycle information.
The reason is that this environment has different game modes, such as Play On, Free Kick, Goal Catch, etc.
The DataExtractor is limited to saving information only in Play On mode because teams usually act differently in other modes.
Hence, around 2000 cycle of each game is in other modes. 

The data of the ball includes the position and velocity of the ball and the pos count of the data.
Also, for each player object, there are position and velocity and the body angle of the player with its pos count.

The order of the features is as follows:
\begin{itemize}
  \item Cycle of the game
  \item Noisy Data
  \begin{itemize}
  \item Ball features(position, velocity, pos count)
  \item Left team players (11 players with position, velocity, body, pos count)
  \item Right team players (11 players with position, velocity, body, pos count)
\end{itemize}
\item Accurate Data
  \begin{itemize}
  \item Ball features(position, velocity, pos count)
  \item Left team players (11 players with position, velocity, body, pos count)
  \item Right team players (11 players with position, velocity, body, pos count)
  \end{itemize}
\end{itemize}

\subsection{Scaling}

Since there are different domains for features, all are scaled within the $[-1,1]$ range.
Regarding the player's position data, where each player's position is defined by X (-52.5 to 52.5) and Y (-34 to 34),  their scaled version is between -1 and 1.
In addition, the maximum speed of the objects in this environment is for the ball with $3m/s$. 
So, we scaled the velocity feature from $[-3,3]$ to $[-1,1]$.
Also, the players' body angle is between -180 and +180 degrees converted to -1 and +1.
Moreover, the maximum pos count is considered $30$ cycles because the agent will remove the player after those cycles pass without seeing a specific player. 
The transformation is from $[0,30]$ to $[0,1]$.
Table \ref{tab_sc} shows each feature's main and scaled domains. 

\begin{table}[]
  \label{tab_sc}
  \centering
  \caption{The comparison of the feature domain before and after the scaling}
  \begin{tabular}{|l|c|c|}
  \hline
  Feature    & \multicolumn{1}{l|}{Domain} & \multicolumn{1}{l|}{Scaled Domain} \\ \hline
  position x & {[}-52.5, +52.5{]}           & {[}-1, +1{]}                        \\ \hline
  position y & {[}-34, +34{]}              & {[}-1, +1{]}                        \\ \hline
  velocity x & {[}-3, +3{]}                & {[}-1, +1{]}                        \\ \hline
  velocity y & {[}-3, +3{]}                & {[}-1, +1{]}                        \\ \hline
  body       & {[}-180, +180{]}            & {[}-1, +1{]}                        \\ \hline
  pos count  & {[}0, +20{]}                 & {[}0, +1{]}                          \\ \hline
  \end{tabular}
  \end{table}

\subsubsection{Imputation}
It is possible that several samples do not contain complete information about the state since the agent forgets the players and the ball after $30$ cycles. 
Therefore, those unknown objects are considered to be outside the field with $(-105, -68)$ position equal to $(-2, -2)$ in the scaled domain.
The pos count of the unknown objects is considered 30.

\subsection{Model Selection}

Predicting a player's position in a soccer simulation game can be considered a time-series problem, as each player's state depends on their own and other players' previous states.
In this context, states can be defined as players and the ball's position and velocity in addition to the players' body angle.

\begin{figure}[!ht]
  \centering
  \includegraphics[width=0.5\textwidth]{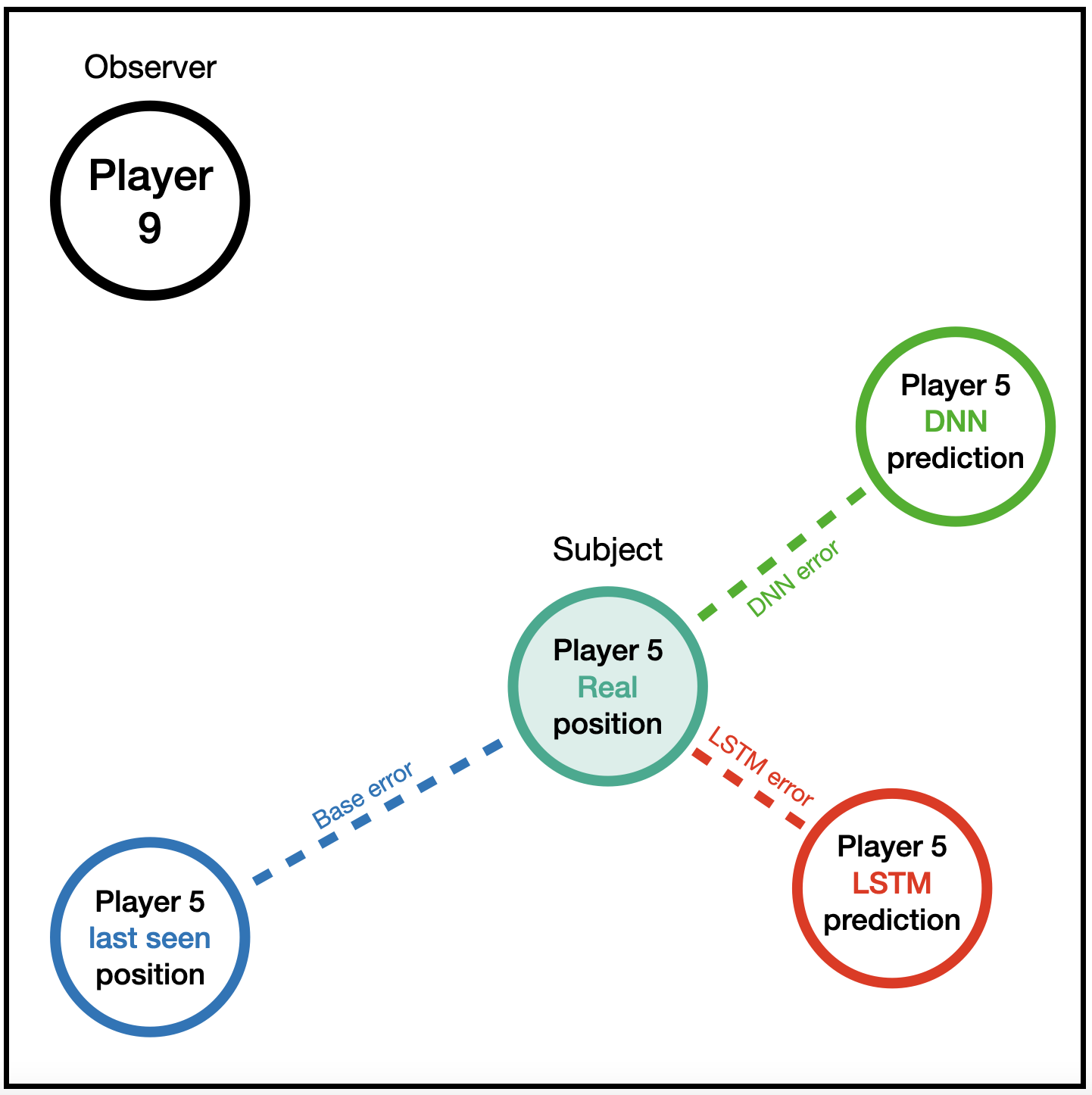}
  \caption{When player 9 sees player 5, the observation contains some noises. Therefore, the blue circle is the position that player 5 is considered, whereas green and red circles are the positions that our methods are predicted.}
  \label{fig_sim}
\end{figure}

In soccer simulation, when a player (observer) does not see another player (subject) for cycles, the observer assumes the subject is still in the last position that it has been seen. 
In the first phase, to have a prediction of the subject's new position (only the opponent's positions), we employed an array of different deep neural network\cite{dnn} architectures to have a prediction baseline. 
However, because our problem can be considered a time-series problem, we gave a possibility that recurrent neural network architectures might be sufficient for this problem.

In phase two, we employed one of the most famous recurrent neural network modules called long short-term memory (LSTM) \cite{lstm} and implemented several other architectures using this module for prediction. 
In both phases, we generated several architectures by changing the number of layers and the number of neurons in each layer to find the optimal network.
Table~\ref{my-table} provides detailed information on the tested architectures. 
In addition, Fig.~\ref{fig_sim} shows the error of different models to understand better.

In phase three, we tweaked the \textbf{lookback period} that indicates the number of prior cycles employed to predict the following cycle.
The best strategy is to test different lookbacks and find the lookback's optimal value. 
For this purpose, we repeated the experiment with different lookbacks (5, 10, and 15) to find the optimal value.

\begin{table}[ht]
  \centering
  \caption{
  A comprehensive overview of neural network architectures implemented (only hidden layers)}
  \label{my-table}
  \begin{tabulary}{\textwidth}{>{\centering\arraybackslash}J>{\centering\arraybackslash}J>{\centering\arraybackslash}J>{\centering\arraybackslash}J>{\centering\arraybackslash}J>{\centering\arraybackslash}J}
  \hline
  \textbf{Model} & \textbf{Layer 1} & \textbf{Layer 2} & \textbf{Layer 3} & \textbf{Layer 4} & \textbf{Layer 5}\\
  \hline
  1 (Phase 1) & 128 & 64 & - & - & -\\
  2 (Phase 1) & 256 & 128 & - & - & -\\
  3 (Phase 1) & 512 & 256 & - & - & -\\
  4 (Phase 1) & 256 & 128 & 64 & 32 & -\\
  5 (Phase 1) & 512 & 256 & 128 & 64 & 32\\
  \hline
  6 (Phase 2) & 256 (LSTM) & 128 & - & - & -\\
  7 (Phase 2) & 512 (LSTM) & 256 & - & - & -\\
  8 (Phase 2) & 128 (LSTM) & 64 & 32 & - & -\\
  9 (Phase 2) & 512 (LSTM) & 256 & 128 & 32 & -\\
  \end{tabulary}
  \end{table}
  
\begin{figure}[!ht]
  \centering
  \includegraphics[width=0.8\textwidth]{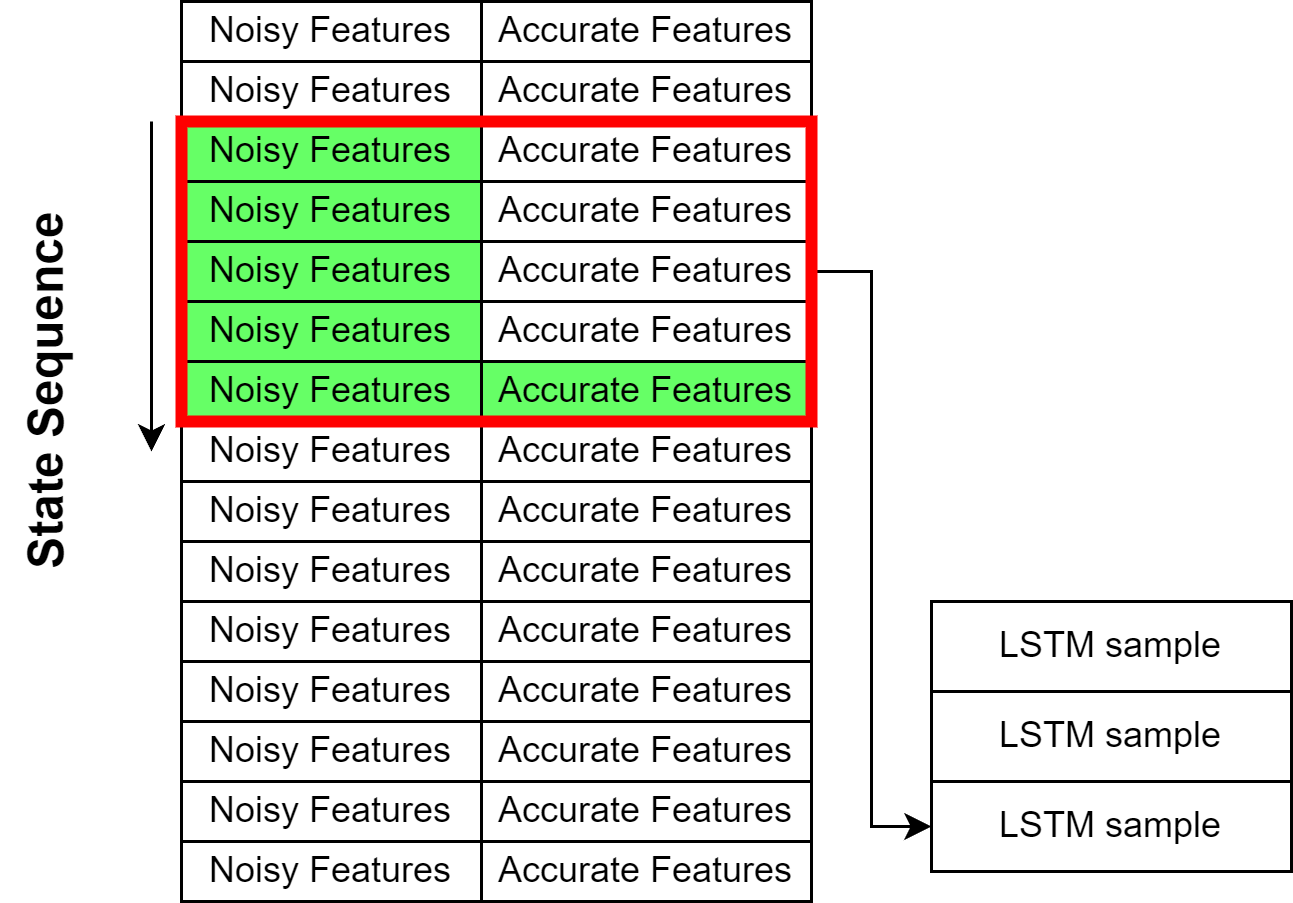}
  \caption{The red window is a window frame that moves down one cycle each time and saves a new LSTM sample that contains green cells.}
  \label{lstmsample}
\end{figure}

For LSTMs with different lookback periods, we used the same dataset.
Sequences of states mean the start and end cycle of each Play On mode in a game.
The short sequences that are less than the lookback period are removed since they are not used.
After that, A window frame with the size of the lookback period is defined to extract the data from the dataset. 
As Fig.~\ref{lstmsample} shows, one complete sample of data contains the noisy data of objects in the whole data frame and the exact position of 11 opponents in the last cycle of the window frame.
The window frame moves through all the state sequences cycle by cycle and makes data from all the games.
The size of the state sequence in this Fig.~\ref{lstmsample} is 15, which is enough for the lookback period.
The state sequence contains both Noisy and Accurate features, where the Noisy feature is for the input of the LSTM model, and the Accurate feature is for the model's output.

After training each of the models, we performed a test procedure. 
In the test phase, new data that is not used in the training phase are given to the models. After that, the predictions of positions are scaled to their main domain. In addition, We consider player 5 of the opponent team to compare the accuracy of different methods. Each model's prediction in test samples is compared to the real position of player 5.

\section{Experiments}

For the experiment, we repeated training and evaluation with different deep neural network architectures and LSTM-based architectures with different lookback periods.





In addition to the lookback parameter, two parameters affect model performance considerably. 
The first one is the player distance(observer distance from the subject).
More distance between the observer and the subject causes more noise to the observer's estimation of this distance, and as a result, this contributes to the error increase.
The other and more important parameter is the pos count.
Of course, as time passes, it becomes more probable that the subject gets further from the last position seen by the observer, and the base error increases.

\begin{figure}[!ht]
  \centering
  \includegraphics[width=1\textwidth]{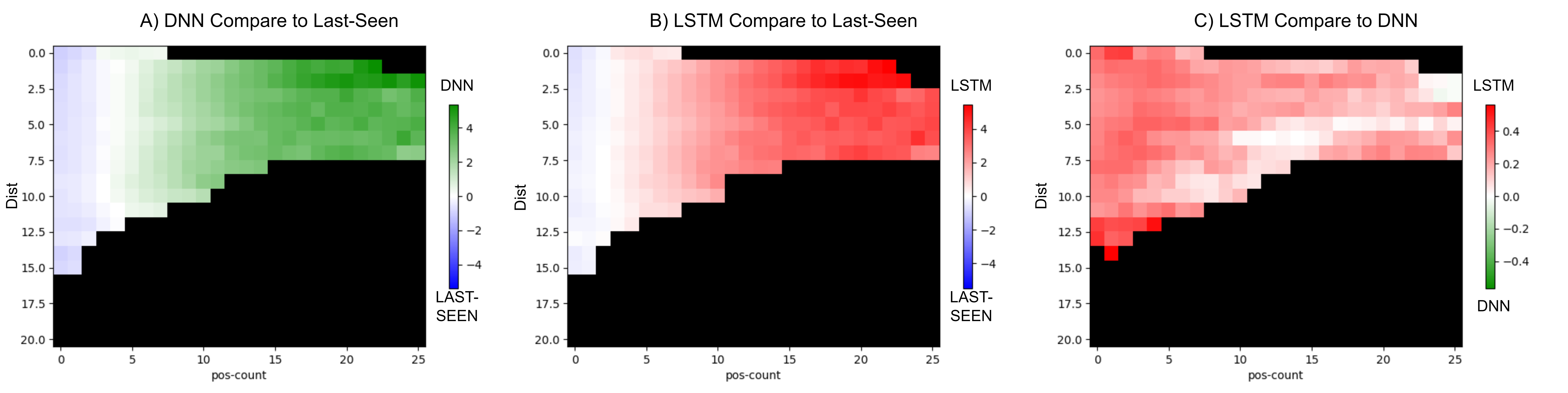}
  \caption{The figures show the comparison of the models with the last-seen method. The colors show the performance of the methods. Green is for DNN, red is for LSTM, and blue is for the last-seen method.}
  \label{heatmaps}
\end{figure}

To better understand each model's performance and a more comprehensive comparison, we employed two parameters of pos-count and player's distance and another parameter of err\_sub (subtraction of errors of two models under comparison).
Fig~\ref{heatmaps} shows the three comparisons.
Each heatmap shows the numerical difference between the distance of the two methods w.r.t. the accurate position. 
The positive number illustrates that the first method's performance was better than the second method, whereas the negative number shows the opposite.

\begin{figure}[!ht]
  \centering
  \includegraphics[width=0.8\textwidth]{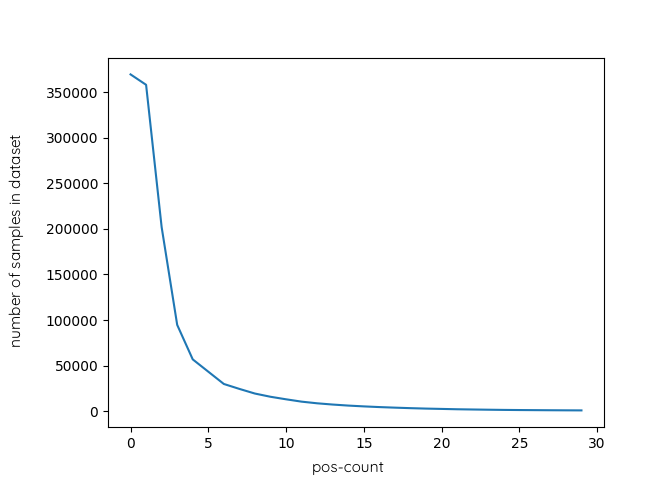}
  \caption{This figure shows the number of available samples in the dataset based on pos count}
  \label{fig_data_pc}
\end{figure}

From Fig.~\ref{heatmaps}, we can observe that as the pos-count increases, the difference between the last-seen (base) prediction and ML prediction increases, which indicates the superior performance of the DNN and LSTM in comparison with the last-seen (base) prediction.
Moreover, based on the \ref{heatmaps} C) the LSTM model performs significantly better than the DNN model.
The black blocks on the heatmap are normal missing data.
There is not enough information for high pos-counts because, after a few cycles, the observer sees the subject again.
Fig.~\ref{fig_data_pc} shows the number of samples available in the dataset for each pos-count. 

\begin{figure}[!ht]
  \centering
  \includegraphics[width=1\textwidth]{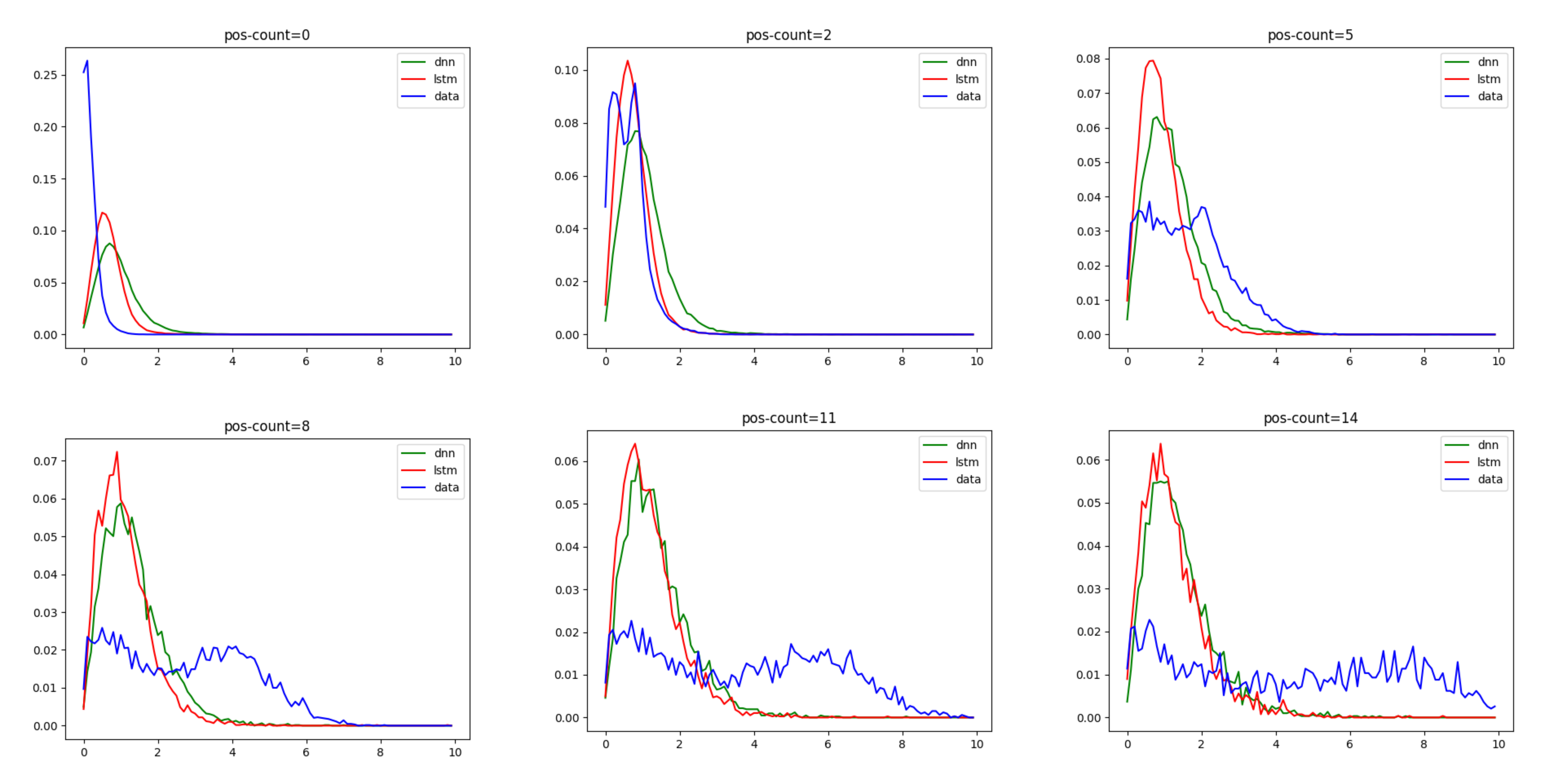}
  \caption{The Comparison of distance rate between 3 methods last-seen(labeled as data), lstm, and dnn methods. The Y axis is the rate of distance error, and the X axis is the distance.}
  \label{fig_pc}
\end{figure}

Fig.~\ref{fig_pc} demonstrates the rate of distance error in different pos counts. 
As the figure suggests, the best situation to use this model is when the pos count of an opponent player is more than or equal to 3, and the distance between the observer and the opponent is more than $1.5m$. 


We compared all of the models with each other, as the results presented in the heatmaps (as shown in  Fig.~\ref{heatmaps}).
The best models that perform the highest accurate prediction are mentioned below.

\begin{itemize}
    \item \textbf{LSTM} with 512 and 256 neurons and all ReLU activations.
    \item \textbf{DNN} with 512, 256, 128, 64, and 32 hidden neurons all ReLU activations with lookback period $5$ cycles.
\end{itemize}

The LSTM model in the mentioned architecture has the best accuracy among all (DNN and LSTMS models) with a five-year lookback period.
However, the DNN can perform a good prediction based only on one cycle but with lower accuracy than LSTM.

\section{Conclusion}

Soccer Simulation 2D is a league in the RoboCup organization.
This environment has different visual challenges, such as noise and partial observation. 
Several researchers have explored tackle inaccuracy in players' position estimations with different methods, leading to higher performance of the player by increasing the percentage of the successful pass and the ball possession rate in a game.

This study demonstrated the importance of predicting players' positions in the Soccer Simulation 2D league.
Despite previous attempts in this domain, studies have yet to utilize machine learning algorithms to make position predictions.
Our research explored various machine learning algorithms, utilizing both deep neural networks and LSTM-based networks, to predict player positions.
Our proposed model outperformed the current base prediction methods and demonstrated the superiority of LSTM-based networks over deep neural networks.

The results of this study suggest that utilizing recurrent neural networks for position prediction in soccer simulation leagues can offer significant advantages to a team's performance.
Although our model has shown promising results, implementing more advanced networks, such as transformers, can further improve the accuracy of position prediction.
By continuing to explore and refine these techniques, we can unlock the full potential of machine learning in enhancing simulated soccer strategies and performance.

%
%
%

\end{document}